# Green-Blue Stripe Pattern for Range Sensing from a Single Image

Changsoo Je * Kyuhyoung Choi Sang Wook Lee

Sogang University, Seoul, Korea

* Corresponding author: vision@sogang.ac.kr

**Abstract**

In this paper, we present a novel method for rapid high-resolution range sensing using green-blue stripe pattern. We use green and blue for designing high-frequency stripe projection pattern. For accurate and reliable range recovery, we identify the stripe patterns by our color-stripe segmentation and unwrapping algorithms. The experimental result for a naked human face shows the effectiveness of our method.

## 1. Introduction

Three-dimensional (3D) geometric information is what people often try to extract to know about physical objects and to utilize them for a variety of purposes. For this reason, numerous 3D range sensing methods have been proposed. Among them, active sensing methods based on structured lights and cameras are widely used since they typically provide satisfactory accuracy and reliability.

The goal of our research is to develop a reliable method for high-resolution range imaging of dynamic objects using a single video frame for real-time capture. 3D reconstruction of dynamic objects is highly challenging in that only a single projection of structured light pattern should be used for real-time capture.

A number of active sensing systems were designed for high-resolution 3D ranging or for high-speed scanning. The laser stripe-based system, one of the simple active systems, can estimate depths with high resolution using sensor-light calibration and triangulation. However, it requires mechanical motion of the light for object scanning, and this prohibits real-time sensing in most cases. Many coded structured-light patterns are free from mechanical scanning. The well-established structured-light sensing uses multiple projections of binary-encoded light stripes, but this requirement of multiple projections also makes real-time imaging infeasible. The depth resolution is dependent on the number of binary stripe patterns, i.e., the number of projections.

Color can be used to decrease the number of projections, and a color-encoded discrete-stripe structured light method was proposed [Boyer and Kak 1987]. On the other hand, phase-shifted three-cosine fringe method [Huang et al. 1999] and rainbow range finder (RRF) [Tajima and Iwakawa 1990] used color continuous patterns. An attempt was made to minimize the number of color patterns, and a detailed method is given [Caspi et al. 1998]. However, in spite of much research on color-encoded structured light, high-resolution real-time ranging has been hindered by the problems in color-resolution and colorimetric calibration.

In addition, color is difficult to be treated appropriately in structured-light ranging while it is good for increasing data resolution. These difficulties made some previous methods limited to gray or KW (black and white). A method using gray-level ratio for real-time ranging was proposed for obtaining rather crude data [Miyasaka et al. 2000]. A new KW stripe pattern method for real-time ranging was proposed [Hall-Holt and Rusinkiewicz 2001], but it is appropriate only for slowly moving rigid bodies but not for deforming objects.

In this paper, we present our research on rapid high-resolution range imaging based on green-blue stripe pattern. The focus of our research is to investigate accurate and convenient color-stripe identification, and reliable unwrapping. This paper is an updated version of a conference paper [Je et al. 2003].

The rest of this paper is organized as follows. Section 2 describes the design of green-blue stripe pattern for accurate 3D data acquisition, and Section 3 presents a color-stripe segmentation algorithm. Section 4 presents an unwrapping algorithm to remove discontinuity due to the repetitive use of green and blue stripes. In Section 5, the experimental result of a human face is provided, and Section 6 concludes the paper.

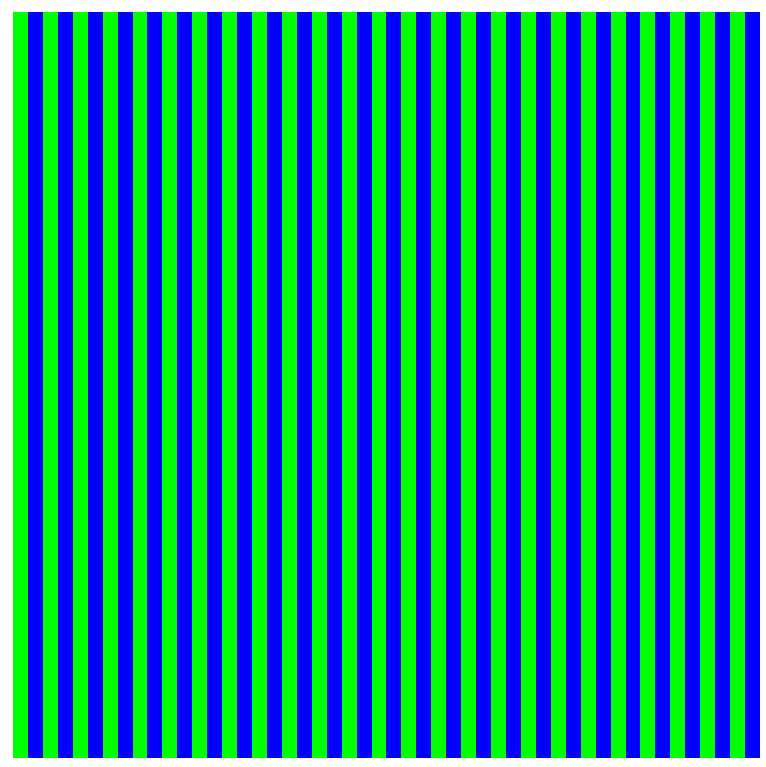

Figure 1. The green-blue stripe pattern image.

## 2. Green-Blue Stripe Pattern Design

We assigned maximum intensity difference for encoding the stripe boundaries in our structured-light pattern to maximize the reliability of stripe segmentation. Only two primary colors, green (0, 255, 0) and blue (0, 0, 255) are used in creating the stripe pattern since the measured dynamic range of red channel is too small in our camera response to projector. The pair of green and blue stripes is repeated in the pattern image for widening the projection area of the light pattern and increasing the data resolution. Figure 1 shows a small part of the green-blue stripe pattern image. The pattern image is projected onto object by a DLP projector (see Figure 2).

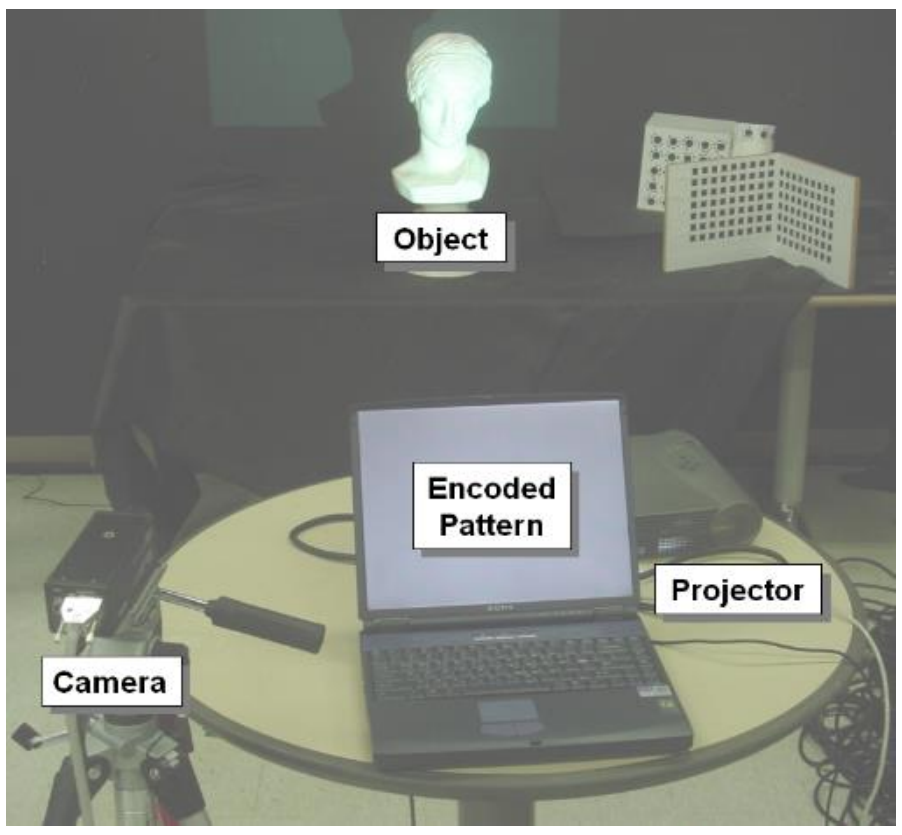

Figure 2. The setup for capturing images of the scene in which the encoded pattern is projected onto an object.

## 3. Color-Stripe Segmentation

The colors of the stripe in the captured scene image need to be reliably recognized (color-stripe segmentation) before the repetition of green-blue stripes is distinguished (unwrapping, see section 4) to identify the light planes built by the projection patterns. Our color-stripe segmentation algorithm consists of motion blurring, color balancing, and thresholding.

### 3.1. Motion Blurring

The green-blue stripe pattern is highly directional since all the stripes in the projection image are mutually parallel. Therefore, the captured images of the scene in which the pattern is projected contain the directionality of the stripe patterns. However, the directionality information in captured images is degenerated by the surface reflectance and noise. We use a motion-blurring filter (directional Gaussian filter) to reduce the degeneration. Consequently, the reliability of stripe colors is improved in most images, and Figure 3 provides comparison of the original and motion-blurred images.

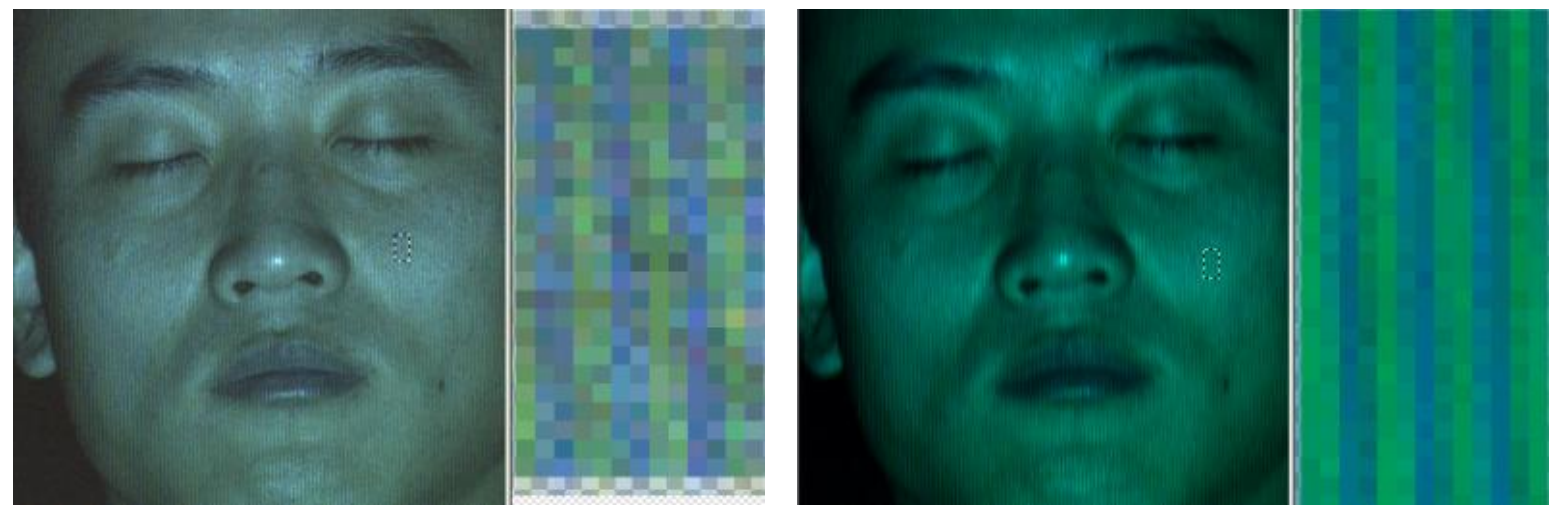

Figure 3. The captured image of a human face and its enlarged small part (left), and the motion-blurred image of the captured image and its enlarged small part (right).

### 3.2. Color Balancing

The camera we use has a white balance function. However the color balance function does not make exact color balance, and cannot resolve local color unbalance due to non-uniform reflectance of object surface. Hence we need a color-balancing step before determination of stripe colors.

First, we compute the mean color in a window of an appropriate size:

$$\bar{C}_i(x, y) = \frac{\sum_{x,y} C_i(x, y)}{\sum_{x,y}}, \quad i = 1, 2, 3. \tag{3.1}$$

where $(C_1, C_2, C_3) \equiv (Red, Green, Blue)$. For color balance, we will use the color defined by the equation,

$$c_i = \frac{C_i}{\bar{C}_i}, \quad i = 1, 2, 3. \tag{3.2}$$

Using $c_i$ is especially useful for range sensing of a human face since a human face has non-uniform reflectance. In case surface color varies moderately with the spatial position as in a human face, the local average values of RGB are recommended for $\bar{C}_i$, and in case surface color is relatively uniform as in a plaster figure, the window size for color balancing may be larger or whole image size.

### 3.3. Local Thresholding

The green-blue stripe pattern has only two intensity codes (green and blue), and thus thresholding is proper for color classification. We use a local threshold rather than a global threshold since reflectance of object surface is not uniform usually. The difference of the results by global and local thresholding is shown in Figure 4.

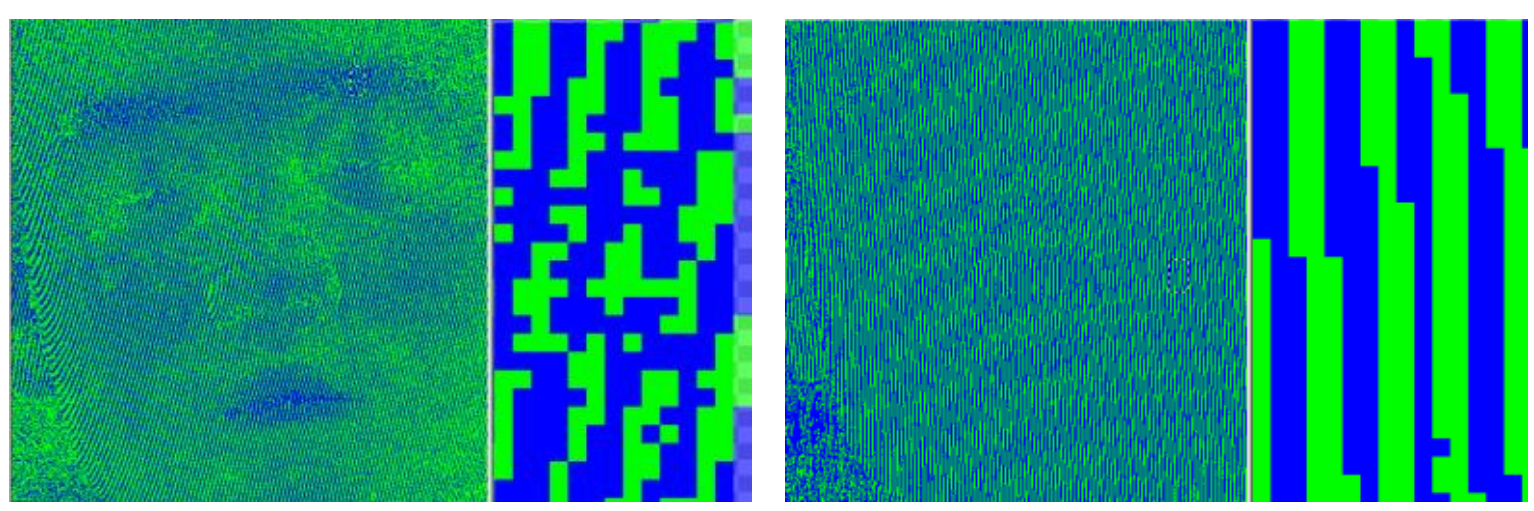

Figure 4. The global threshold-based color-classified image and its enlarged small part (left), and the local threshold-based color-classified image and its enlarged small part (right).

## 4. Unwrapping

The classified color codes need to be unwrapped for removing the color-stripe redundancies. Since the green-blue stripe pattern is of extremely high frequency, unwrapping is quite difficult. We assume the color codes of most pixels in the captured image are correctly classified. Our unwrapping algorithm is as follows (see Figure 5).

**------------------------------- Unwrapping Algorithm -------------------------------**

1. Unroll each row by transition through the row

2. Unroll every row by relation with upper neighbor row

3. Determine believable Rows

4. Unroll each pixel by relation with upper neighbor pixel in believable rows

5. Unroll each row by transition through the row

6. Unroll every row by relation with upper neighbor pixel

7. Unroll each pixel by relation with upper neighbor pixel

**-----------------------------------------------------------------------------------------------**

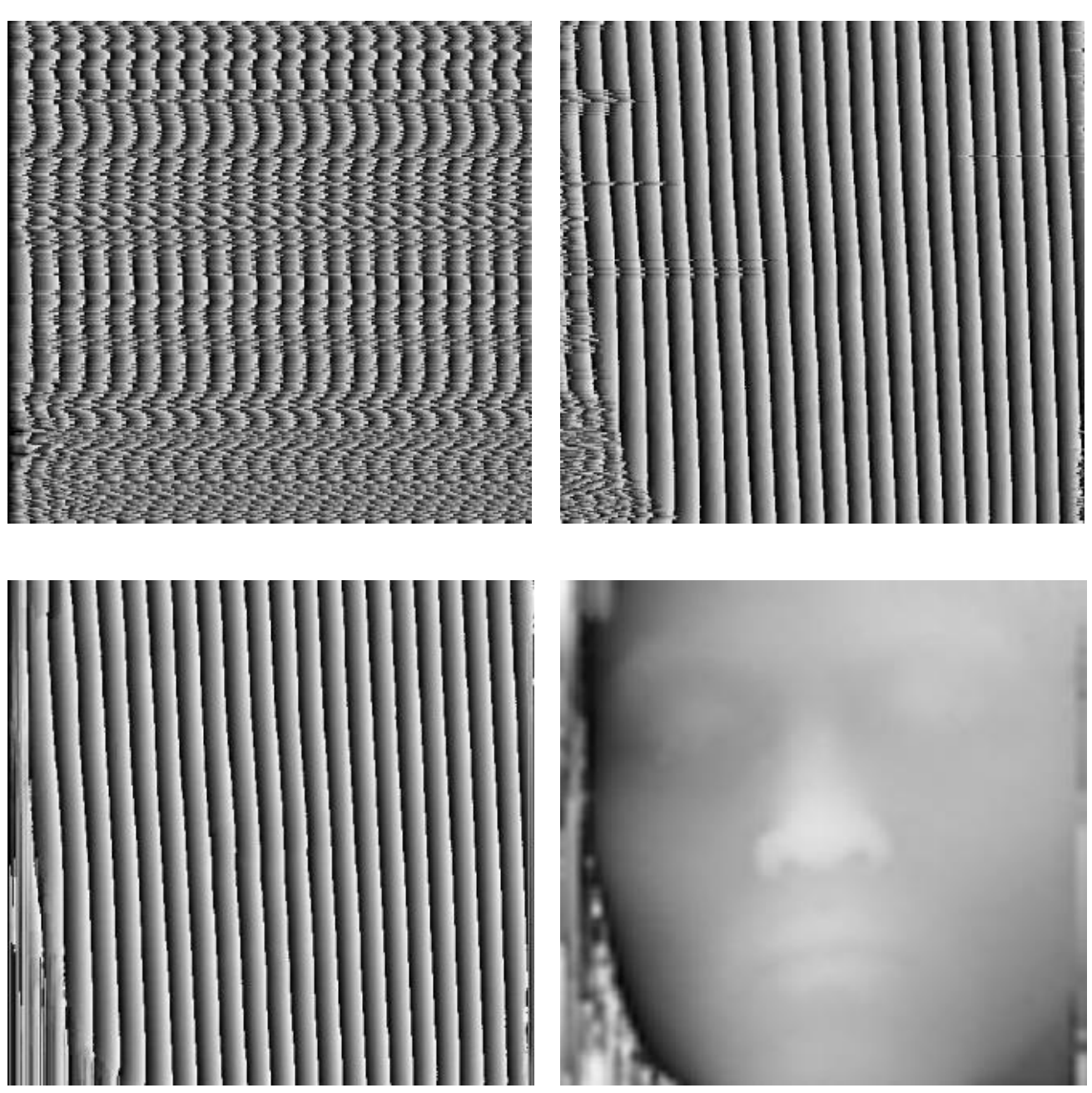

Figure 5. The sequence of the range recovery processing. Unrolling each row by transition through the row (top left). Unrolling every row by relation with upper neighbor row (top right). Correction (bottom left). The range result of a naked human face from a single image (bottom right).

Although some misclassified pixels exist in most images the algorithm produces high-accuracy unwrapping results. The unwrapping algorithm is of low cost (no functional optimization), and has high feasibility (pixel-by-pixel processing), flexibility (the number of reference pixels and that of reference rows are changeable), and robustness to misclassified pixels and the high frequency of the stripe pattern.

## 5. Experimental Result

The experimental setup consists of an Infocus DLP 1024×768 color projector, a Sony XC-003 640×480 3-CCD camera, and Pentium III computers for projecting, capturing and processing. Figure 2 shows the setup for experiment. In Figures 3-5, the range result is obtained from only a single image of a naked human face, and the result shows the effectiveness of our method.

## 6. Conclusion and Future Work

We presented a novel method for rapid high-resolution range sensing using green-blue stripe pattern. We use green and blue for designing high-frequency stripe projection pattern. For accurate and reliable range recovery, we identify the stripes using our color-stripe classification and unwrapping algorithms. For robust color classification, the color-stripe segmentation algorithm performs motion blurring, color balancing, and local thresholding. The unwrapping algorithm resolves the high-frequency color-stripe redundancy efficiently and reliably. The experimental result from a single image of a naked human face showed the effectiveness of the method. For future work, we are interested in applying the presented segmentation and unwrapping algorithms to color-stripe permutation pattern [Je et al. 2012] for accurate range acquisition.